\documentclass[times,twocolumn,final,authoryear]{elsarticle}

\usepackage{ycviu}
\usepackage{framed,multirow}

\usepackage[colorlinks,
linkcolor=red,
anchorcolor=blue,
citecolor=green
]{hyperref}

\usepackage{amssymb}
\usepackage{latexsym}

\usepackage{url}
\usepackage{xcolor}
\definecolor{newcolor}{rgb}{.8,.349,.1}

\usepackage{times}
\usepackage{epsfig}
\usepackage{graphicx}
\usepackage{amsmath}
\usepackage{amssymb}
\usepackage{multirow}
\usepackage{subfig}
\usepackage[export]{adjustbox}
\usepackage{enumitem}

\newcommand{\warn}[1]{\textcolor{black}{#1}}

\journal{Computer Vision and Image Understanding}

\begin{document}

\thispagestyle{empty}

\begin{frontmatter}

\title{MTRNet++: One-stage Mask-based Scene Text Eraser}

\author[1]{Osman \snm{Tursun}\corref{cor1}} 
\cortext[cor1]{Corresponding author}
\ead{w.tuerxun@qut.edu.au}
\author[1]{Simon \snm{Denman}}
\author[1,2]{Rui \snm{Zeng}}
\author[1]{Sabesan \snm{Sivapalan}}
\author[1]{Sridha \snm{Sridharan}}
\author[1]{Clinton \snm{Fookes}}

\address[1]{Signal Processing, Artificial Intelligence and Vision Technologies (SAIVT), Queensland University of Technology, Australia}
\address[2]{Brain and Mind Centre, The University of Sydney, Australia}

\received{}
\finalform{}
\accepted{}
\availableonline{}
\communicated{}

\begin{abstract}
A precise, controllable, interpretable and easily trainable text removal approach is necessary for both user-specific and large-scale text removal applications. To achieve this, we propose a one-stage mask-based text inpainting network, MTRNet++. It has a novel architecture that includes mask-refine, coarse-inpainting and fine-inpainting branches, and attention blocks. With this architecture, MTRNet++ can remove text either with or without an external mask. It achieves state-of-the-art results on both the Oxford and SCUT datasets without using external ground-truth masks. The results of ablation studies demonstrate that the proposed multi-branch architecture with attention blocks is effective and essential. It also demonstrates controllability and interpretability.
\end{abstract}


\end{frontmatter}

\section{Introduction}
\label{sec:intro}
Text removal is the task of inpainting text regions in scenes with semantically correct backgrounds. It is useful for privacy protection, image/video editing, and image retrieval \citep{tursun2019component}. Recent studies with advanced deep learning models \citep{zhang2018ensnet,tursun2019mtrnet,nakamura2017scene} explore removing text from real world scenes. Previously, text removal with traditional methods has only been effective for text in fixed positions with a uniform background from digital-born content.

\let\thefootnote\relax\footnotetext{*** Under CVIU review. ***}

\begin{figure}[!t]
	\centering
	\subfloat[]{
	\label{fig:ca}
	\includegraphics[width=0.22\linewidth]{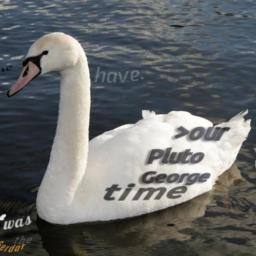}
	\includegraphics[width=0.22\linewidth]{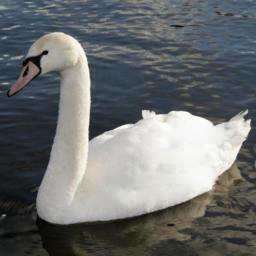}}
	\subfloat[]{	
	\label{fig:cb}
	\includegraphics[width=0.22\linewidth]{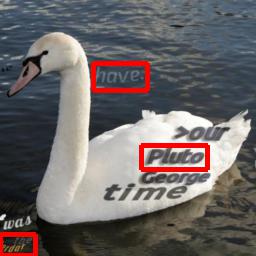}
	\includegraphics[width=0.22\linewidth]{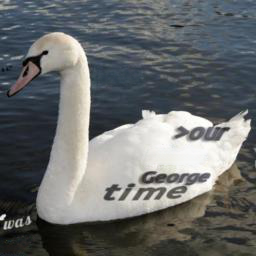}}

	\subfloat[]{
	\label{fig:cc}	
	\includegraphics[width=0.22\linewidth]{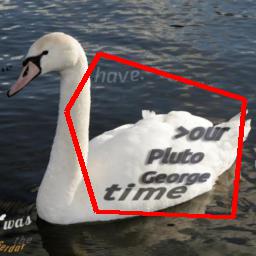}
	\includegraphics[width=0.22\linewidth]{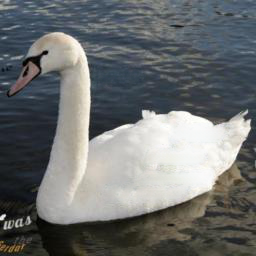}}
	\subfloat[]{
	\label{fig:cd}	
	\includegraphics[width=0.22\linewidth]{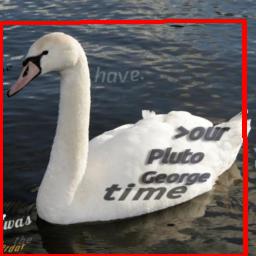}
	\includegraphics[width=0.22\linewidth]{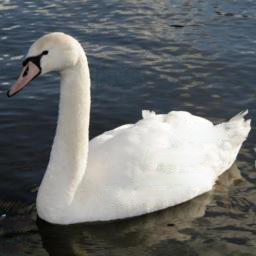}}\textbf{}
	
	\caption{MTRNet++ is a one-stage text removal approach which removes text either with or without a mask. In the without mask case, MTRNet++ will remove all text in a scene as shown in Fig. \ref{fig:ca}. While with a mask, MTRNet++ is able to only remove text under the masked region as shown in Figs. 1b-d. Note that masked regions are labelled with polygons using red outlines.}
	\label{fig:concept}
\end{figure}

Text removal is a challenging task as it inherits the challenges of both text-detection and inpainting tasks. \cite{zhang2018ensnet} claims that text removal requires stroke-level text localization, which is a harder and less studied topic compared to bounding-box-level scene text detection \citep{gupta2016synthetic}. On the other hand, realistic inpainting requires replacing/filling unwanted objects in scenes with perceptually plausible content. Deep learning approaches generate perceptually plausible content by learning the distribution of training data. Text can be placed in any region or on any object, and as such a text inpainting model is required to learn a wide range of distributions, exacerbating the challenge.

Recent text removal studies are based around two main paradigms: one-stage and two-stage methods. The one-stage approach uses an end-to-end model, which is free of auxiliary inputs; however, its generality, flexibility and interpretability are limited. EnsNet \citep{zhang2018ensnet} is the state-of-the-art one-stage text removal method. It removes all text that is written in scripts on which it was trained. Although it supports selective text removal with some extra steps (cropping or overlapping), this introduces new issues such as color discontinuities or loss of context. Moreover, it lacks interpretability that makes fixing failure cases troublesome. Finally, yet importantly, we found that a one-stage approach without an explicit text localisation mechanism fails to converge on a large-scale dataset within a few epochs. For example, Pix2Pix \citep{isola2017image} and EnsNet do not converge on a large-scale dataset with the same amount of training as their counterparts did according to the results reported in MTRNet \citep{tursun2019mtrnet} and this work.
	


\warn{The two-stage approach decomposes the text removal task into text detection and text inpainting sub-problems. Text detection can be manual or automatic. MTRNet (Tursun et al., 2019b), for example, is a representative inpainting stage of the two-stage approach with a deep neural network.} The main advantage of a two-stage approach is that it has an awareness of the text regions that require inpainting. With explicit text regions, it gains generality and controllability. Two stage approaches can be effortlessly adapted to remove text in various scripts, and are able to remove or keep text based on selection. Such methods have strong interpretability as well; for example it is easier to understand if failure cases are caused by inaccurate detection or poor inpainting. Despite the promise of two-stage methods, they are limited in that they rely on a text localisation front-end. Compared to a one-stage approach, two stage approaches are inefficient and have a complex training process as at least two networks need to be trained.



In this work, we propose a one-stage text removal approach, MTRNet++. It can remove text either with a mask as per MTRNet, or without a mask as per EnsNet as shown in Fig. \ref{fig:concept}. It inherits the idea of MTRNet of using a text region mask as an auxiliary input. However, the network has a different architecture to MTRNet, and is composed of \textit{mask-refine}, \textit{coarse-inpainting}, and \textit{fine-inpainting} branches. The mask-refine and coarse-inpainting branches generate intermediate results, while the fine-inpainting branch generates the final refined results. Moreover, compared to MTRNet, it can remove text under very coarse masks as shown in Figs. \ref{fig:cc} and \ref{fig:cd}.

The mask-refine branch is designed to refine a coarse mask into an accurate pixel-level mask. The mask-refining branch is introduced for the following reasons: (1) To remove text as EnsNet does without requiring third-party text-localisation information, but also to allow utilisation of third-party information as done by MTRNet. (2) To provide interpretability, flexibility and a generalisation ability. (3) To provide attention scores to the coarse-inpainting branch. (4) To speed up the convergence of the network on a large-scale dataset. 

The coarse-inpainting branch is a parallel branch to the mask-refine branch, which performs a coarse inpainting using the coarse mask. The coarse-to-fine framework has been shown to be beneficial for realistic inpainting \citep{yu2018generative, ma2019coarse}, and the coarse-branch is guided by the attention scores generated by the attention blocks that map intermediate features of the mask-refine branch to weights. 

The fine-inpainting branch is introduced for refining the results of the coarse-inpainting branch with precise masks from the mask-refine branch. The coarse results are blurry and lack details. The fine-inpainting branch increases inpainting quality. In this work, for efficiency, a light sub-network is used as the fine-inpainting branch.

In summary,  our contributions are as follows:


\begin{itemize}
	\item We propose a novel one-stage architecture for text removal. With this architecture, MTRNet++ is free from external text localisation methods, yet can also leverage external information.
	
	\item MTRNet++ achieves state-of-the-art quantitative and qualitative results both on Oxford \citep{gupta2016synthetic} and SCUT \citep{zhang2018ensnet} datasets without external masks. Ablation studies shows the proposed architecture and its components play important roles.
	
	\item MTRNet++ is a fully end-to-end trainable network and is easily trainable.
	It converges on large-scale datasets within an epoch. It also demonstrates controllability and interpretability.
\end{itemize}

We also introduce other incremental modifications regarding training losses, training strategy and the discriminator, which will be discussed in Section \ref{sec:met}.

The rest of the paper is organised as follows. The next section presents related literature. Section 3 illustrates the proposed network architecture (generator and discriminator), training losses and training strategy. Section 4 presents experiments, ablation studies and analysis. Finally, we provide a brief summarisation of our work in Section 5.
\section{Literature}
\label{sec:lit}

Text removal is a special case of image inpainting, which usually requires the assistance of a text-detection method for text localization. Early text removal approaches \citep{khodadadi2012text,wagh2015text,tursun2019component} are two-stage methods based on either traditional text detection or inpainting approaches. With the advance of deep learning, many classical problems including image inpainting are solved in a single stage using a deep encoder-decoder neural network \citep{mao2016image}. By being fed numerous examples, the neural network learns to map the input to the text-free ground-truth. \cite{nakamura2017scene} proposed the first one-stage the text removal approach, which is a patch-based skip-connected auto-encoder. They trained the network by providing patches with text and patches without text. Inpainting results of the early one-stage approach are blurry and lack detail, as networks are trained with only pixel intensity-based losses.

Later works improve inpainting via new losses generated from a neural network \citep{yang2017high,liu2018image,isola2017image,iizuka2017globally}, such as content and style losses which are calculated using classification networks pre-trained on image-net. Adversarial losses obtained from a generative adversarial network (GAN) \cite{yu2018generative} have also shown promise for inpainting. Recent works \citep{jo2019sc,nazeri2019edgeconnect} show that using both a pre-trained classification network and a newly trained discriminator for image inpainting is beneficial. \cite{zhang2018ensnet} trained a one-stage text removal network with both a pre-trained and adversarial network, achieving promising results at a higher resolution.

Inpainting has also been improved by introducing new types of convolution \citep{liu2018image,yu2018free,iizuka2017globally} and multi-stage architectures \citep{yu2018free,nazeri2019edgeconnect}. Inspired by these and building upon the work of MTRNet \citep{tursun2019mtrnet}, we propose a mask-based one-stage approach for text removal.






%
%

\section{MTRNet++}
\label{sec:met}

MTRNet++ is text-inpainting network, and is formulated as a conditional generative adversarial network \cite{isola2017image}. It is composed of a multi-branch generator $G$ and a discriminator $D$. In the following sections, we illustrate their structures and training strategy.

\subsection{Multi-branch Generator}
\begin{figure*}[t!]
	\centering
	\includegraphics[width=\textwidth]{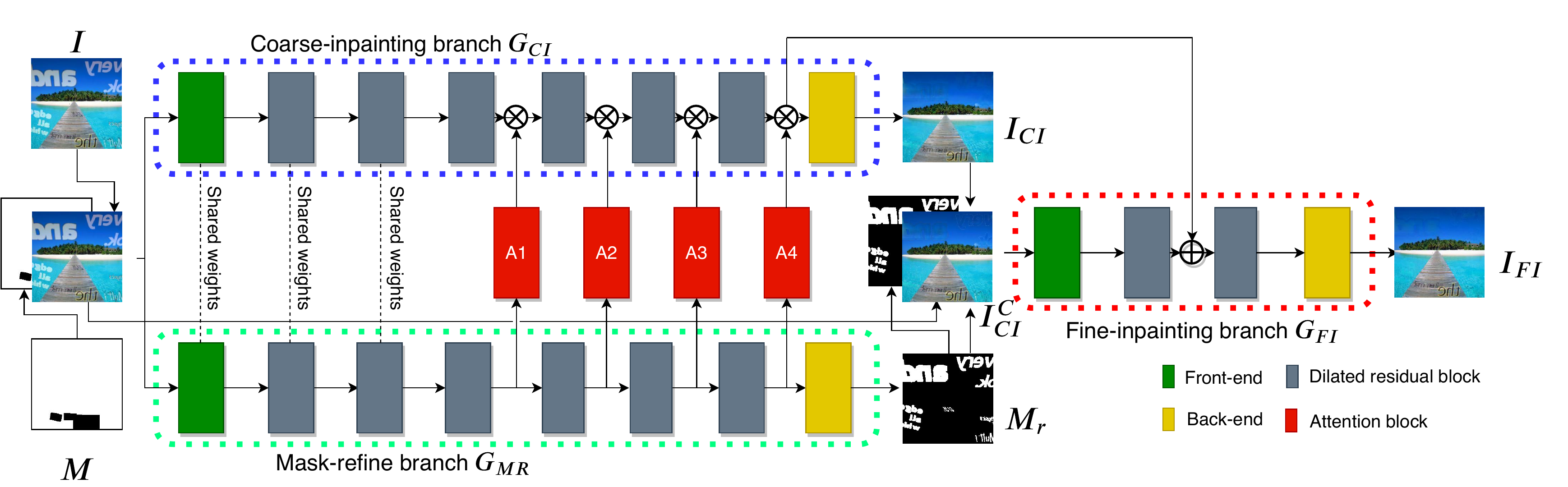}
	\caption{The overall structure of the proposed one-stage text removal method, MTRNet++. It is composed of mask-refine $G_{MR}$, coarse-inpainting $G_{CI}$ and fine-inpainting $G_{FI}$ branches. An input to all branches is a channel-wise concatenation of an image and a mask. \warn{Note that while a mask can be supplied to the network, this is not mandatory and an empty mask where no regions are excluded can be provided. In this case, the network will determine the text regions automatically and supress them.}}
	\label{fig:model}
\end{figure*}
\noindent

The proposed method is a multi-branch, one-stage text removal approach. As shown in Fig. \ref{fig:model}, its generator, $G$, is composed of three branches: mask-refine $G_{MR}$, coarse-inpainting $G_{CI}$, and fine-inpainting $G_{FI}$ branches. All branches have a similar architecture inspired by the network proposed by  \citep{johnson2016perceptual}. Each branch has a front-end, a mid-section, and a back-end. The front-end is composed of down-sampling convolutional layers that reduce the image dimensions by a factor of two, while the back-end is a transposed version of the front-end. The mid-section is composed of a set of residual blocks \citep{he2016deep}. $G_{CI}$ and $G_{MR}$ each have six residual blocks. Their first two residual blocks share weights, while there are attention blocks connecting the last-four residual blocks. As per \cite{nazeri2019edgeconnect}, dilated convolutions \citep{YuKoltun2016} with a dilation factor of two are applied in all residual blocks. MTRNet++, therefore, has a larger receptive field compared to MTRNet, which is important when inpainting text with a large font size. The detailed structures of the aforementioned components are described in \ref{ap:ap1}.


\noindent{\bfseries{Mask-refine}}
The mask refining process transforms the coarse mask, $\mathbf M$, to a refined mask, $\mathbf M_{r}$, that only covers text pixels. 
As shown in Fig. \ref{fig:model}, with a concatenation of the input image $\mathbf I$ and $\mathbf M$ as input, $G$'s mask refine branch, $G_{MR}$, generates the refined mask,
\begin{equation}
\mathbf M_r =  G_{MR}(\mathbf I, \mathbf M).
\end{equation}
 
The objective is to keep dilating a coarse mask such that it covers all text pixels under it. In other words, high recall is expected instead of high precision. To boost the recall, Tversky loss \citep{salehi2017tversky}, presented in Eq. \ref{eq:tversky}, is introduced as this offers a better trade-off between precision and recall by controlling the weights of false positives and false negatives, which are $\alpha$ and $\beta$, in Eq. \ref{eq:tversky}. In experiments, $\alpha$ and $\beta$ are set to 0.1 and 0.9.

\begin{equation}
\mathcal{L}_{MR} = \frac{\alpha \left \|\mathbf M_r \odot \mathbf M^c_{gt} \right \|_1  + \beta \|\mathbf M_r^c \odot \mathbf M_{gt}\|_1}{\|\mathbf M_r \odot \mathbf M_{gt}\|_1 + \alpha \|\mathbf M_r \odot \mathbf M^c_{gt}\|_1   + \beta \|\mathbf M_r^c \odot \mathbf M_{gt}\|_1},
\label{eq:tversky}
\end{equation}
where the superscript ``c'' indicates the  inverse of an image, and $\odot$ denotes the Hadamard product.

Data augmentation is applied to the masks during training. We apply two augmentation techniques: \textit{padding} and \textit{filtering}. $\mathbf M$ is padded with a random padding factor which is smaller than half of the biggest side of $\mathbf M$. The notation $\mathbf M_{p=n}$ represents the padded $\mathbf M$ with a padding size $n$. $\mathbf M_{p=0}$ therefore represents the mask generated with ground-truth word/character level bounding box labels. In training, $10\%$ of masks are padded with biggest side of the mask to ensure the model can adapt to the no mask scenario. With random padding, the generator becomes invariant to mask scale. For filtering, 20$\%$ of bounding boxes are filtered to keep some text from being removed. Via filtering, generators become aware of mask position and meaning. Note that padding is applied after filtering.



%
%

\noindent{\bfseries{Coarse-inpainting}} $G_{CI}$'s coarse-inpainting branch generates a low quality inpainted image,
\begin{equation}
\mathbf I_{CI} =  G_{CI}(\mathbf I, \mathbf M_{p=n}),
\end{equation}

\warn{which is used to create a coarse composited image $\mathbf I_{CI}^C$ by merging with $\mathbf I$ based on $\mathbf M_r$ as follows},

\begin{equation}
\mathbf I_{CI}^C =  \mathbf M_r \odot \mathbf I_{CI} + (\mathbf 1 - \mathbf M_r) \odot \mathbf I.
\end{equation}

\noindent{\bfseries{Attention blocks}} Both the mask-refine and coarse-inpainting tasks are focussing on text which needs to be inpainted. $G_{MR}$ is explicitly trained with ground-truth text labels, thus it captures more accurate text position compared to $G_{CI}$. Here, four attention blocks are applied to transfer text position information to $G_{CI}$ from $G_{MR}$.

\noindent{\bfseries{Fine-inpainting}}
$G_{FI}$ utilises both $\mathbf I_{CI}^C$ and $\mathbf M_{r}$ as additional auxiliary inputs to generate
\begin{equation}
\mathbf I_{FI} = G_{FI}(\mathbf I_{CI}^C, \mathbf M_r), 
\end{equation}
which is more precise in-painted image and has a more realistic structure in the inpainted area. In the training stage, $M_{r}$ is randomly replaced by $M_{gt}$. The probability of this replacement decreases from 1 till 0.1 every 2,000 steps in increments of 0.1. We introduce this progressive replacement as the accuracy of $M_{r}$ is very low initially and that can impact the convergence of the branch $G_{FI}$.

\noindent{\bfseries{Training losses}} Both coarse-inpainting and fine-inpainting branches are trained with L1, perceptual, style and adversarial losses. These losses are also used in EnsNet, though we introduce some modifications to these losses.  Note that style and perceptual losses are calculated based on the last activation maps before the pooling layer after the top five convolution blocks of the VGG-19 network \citep{simonyan2014very}.

To increase the inpainting quality of text regions covered by masks, weighted L1 loss, weighted style loss and weighted perceptual loss are applied. Weights are assigned based on the input masks. Feature or outputs under masked regions are multiplied by the weight parameter $\lambda$. In experiments, $\lambda$ is empirically set to 10.  

The weighted L1 loss $\mathcal L1$ is defined as


\begin{equation}
\mathcal L1 = \sum_{\mathbf I\in \{\mathbf I_{CI}, \mathbf I_{RI}\}} \mathbb{E}[\left \| \mathbf I_{gt}\odot\mathbf M_w - \mathbf I \odot\mathbf M_w \right \|_1],
\label{eq:wl1}
\end{equation}
where $M_w = \lambda \mathbf M_{p=0} + \mathbf M_{p=0}^c$.


The weighted perceptual loss $\mathcal{L}_{perc}$ is defined as
\begin{equation}
\mathcal{L}_{perc} = \sum_{\mathbf I\in \{\mathbf I_{CI}, \mathbf I_{RI}\}} \sum_{i} \frac{1}{N_i}\left \| \O_i (\mathbf I_{gt}) - \O_i (\mathbf I ) \right \|_1,
\label{eq:perc}
\end{equation}
where  $\O_i = \o_i \odot M_w$, and the notation $\o_i$ denotes the activations of the $i$th convolution block. $N_i$ is the number of activation in layer $i$.

The weighted style loss is calculated using the Gram matrix of weighted activation maps, $\O_i$, by 
\begin{equation}
\mathcal{L}_{style} = \frac{\sum_{\mathbf I\in \{\mathbf I_{CI}, \mathbf I_{RI}\}} \sum_{i} \left \| \O_i (\mathbf I_{gt})^T \O_i(\mathbf I_{gt}) - \O_i(\mathbf I)^T \O_i (\mathbf I ) \right \| _1}{C_iH_iW_i},
\label{eq:style}
\end{equation}
where $(H_iW_i)\times C_i$ is the shape of $\O_i$.

In this work, the vanilla adversarial loss is replaced by the hinge loss \citep{miyato2018spectral} to enable efficient and stable training.  The generator  adversarial loss, $\mathcal L_{adv_{G}}$, is defined as 


\begin{equation}
\mathcal L_{adv_{G}} = -\sum_{\mathbf I\in \{\mathbf I_{CI}, \mathbf I_{CI} ^C, \mathbf I_{RI}, \mathbf I_{RI}^C \}} \mathbb{E}[D(\mathbf M_{p=0}, \mathbf I)]
\end{equation}

The generator $G$ is trained with the loss $\mathcal{L}_{G}$,

\begin{equation}
\mathcal{L}_{G} = \mathcal{L}_{MR} + \lambda_{L1}\mathcal{L}_{L_1} + \lambda_{perc}\mathcal{L}_{perc} + \lambda_{style}\mathcal{L}_{style} +
\lambda_{Adv_{G}}\mathcal{L}_{Adv_{G}}
\label{eq:G}, 
\end{equation}

where regularisation parameters $\lambda_{L1}$, $\lambda_{perc}$, $\lambda_{style}$, and $\lambda_{adv_{G}}$ are set to 2.5, 0.05, 12.5 and 0.05 in experiments.

\subsection{Discriminator}
The discriminator is a PatchGAN  \citep{isola2017image} as per MTRNet's discriminator. However, to implement the network as a Wasserstein GAN \citep{miyato2018spectral}, spectral normalization is implemented in each layer as per SN-PatchGAN \citep{yu2018free}. To create a strong discriminator, the original mask $\mathbf M_{p=0}$, is concatenated to the input to each of the top four layers to increase attention on the inpainted regions. The detailed structure of the discriminator is provided at \ref{ap:ap1}. The loss for training the discriminator is defined as follow,

\begin{multline}
\mathcal L_{adv_{D}} = \mathbb{E}[max(\mathbf 0, \mathbf 1 - D(\mathbf M_{p=0}, \mathbf I_{gt}))]\\
+\sum_{I\in \{\mathbf I_{CI}, \mathbf I_{CI} ^C, \mathbf I_{RI}, \mathbf I_{RI}^C \}}\mathbb{E}[max(\mathbf 0, \mathbf 1 + D(\mathbf M_{p=0}, \mathbf I))],
\end{multline}

\subsection{Training Setup and Strategy}
\label{sec:ts}
Our training setup is similar to Edge-connect \citep{nazeri2019edgeconnect}. PyTorch is used for implementation. All models are trained using $256 \times 256$ images with a batch size of eight. The Adam optimiser with $\beta_1 = 0$ and $\beta_2 = 0.9$ is used. The initial learning rate of generator is set to $10^{-4}$. It is divided by 10 when the generator loss value stops decreasing. In total, the learning rate is divided by 10 twice. As we used a WGAN, the learning rate of the discriminator is five times that of generator. In large datasets the convergence appears within 200,000 steps, while in small datasets the final convergence appears within 85,000 steps.

\section{Experiment}
\subsection{Datasets and Evaluation Metrics}
In this study, MTRNet++ is primarily compared with the previous state-of-the-art methods, MTRNet and EnsNet. For a fair comparison, the same datasets and evaluation metrics introduced by MTRNet and EnsNet are applied. The comparison includes quantitative and qualitative results. Quantitative results are given for synthetic datasets that have ground-truth, while qualitative visualisations are provided for both synthetic and real datasets. Most of the datasets used for training and evaluation are well-known challenging datasets for the text detection task.

\noindent
{\bfseries{Synthetic Datasets}} The Oxford synthetic real scene text detection \citep{gupta2016synthetic} and SCUT synthetic text removal \citep{zhang2018ensnet} datasets are adapted for training and quantitative evaluation. The Oxford dataset includes around 800,000 images. We randomly selected 95\% images as training images, 10,000 images as test images, and the rest as validation images. In comparison, the SCUT dataset is a small scale dataset. It includes 8,000 training images and 800 test images. The Oxford dataset was initially proposed for real scene text detection. It is chosen for demonstrating the robustness of MTRNet++. In comparison, the SCUT dataset is selected as it was built for text removal and it includes some real cases. It's background images are manually modified images from the ICDAR 2013 \citep{karatzas2013icdar} and ICDAR 2017 MLT \citep{nayef2017icdar2017} datasets.

\noindent
{\bfseries{Synthetic Refined Mask}} MTRNet++ is a mask-based method. It not only uses masks generated by ground-truth word/character-level polygonal bounding boxes, but also requires pixel-level ground-truth masks for the mask-refining task. However, ground-truth refined masks are not provided by Oxford or SCUT datasets. We created ground-truth refined masks based on the pixel value differences between input and ground-truth images. We assume that input and its ground-truth have pixel value differences only on text pixels. To avoid noise, we used a threshold set to 25. Moreover, we also suppress noise in non-text regions based on the word/character level bounding boxes.

\noindent
{\bfseries{Evaluation Metrics}} In previous studies \citep{nakamura2017scene,zhang2018ensnet,tursun2019mtrnet}, text removal methods are evaluated with text detectability and inpainting quality. \cite{nakamura2017scene} assumes that a high quality text removal method will decrease text detectability. To this end, a robust text detection method is applied to text inpainted images, and precision and recall values are calculated. Precision and recall values are expected to be zero after successful text removal. \cite{zhang2018ensnet} introduced evaluation metrics widely used in inpainting to evaluate the quality of the inpainting. They found the quality of the inpainting can't be captured by the precision or recall scores. To evaluate the quality or realistic degree of inpainting, PSNR, SSIM, MSE and MAE scores are calculated when the ground truth images are available. Note that, for text detection, we applied the recent state-of-the-art text detection method CRAFT \citep{baek2019character}, and for evaluation the DetEval \citep{wolf2014evaluation} protocol is used.


\subsection{Outputs of MTRNet++}
MTRNet++'s three branches output a refined-mask, a coarse-inpainted image and a fine-inpainted image for a given input. In this work, these coarse and fine predicted images are referred to as \textit{predicted}. These predicted results are further improved via replacing non-text regions with the corresponding regions in the given input based on the refined-mask. The resultant composited images are referred to as \textit{composited}. We found colour discontinuities appear near the text boundary on composited images. Although they are very subtle discontinuities, text can be recognisable as text. To overcome colour discontinuities, a dilatation operation with a disk of radius seven is applied to the refined-mask before the composition. The dilatation operation not only reshapes the boundaries but also shifts boundaries to regions with fewer colour discontinuities. The composited fine-inpainting results are used when MTRNet++ is compared with other methods as these represent the best output of the network.

\subsection{Comparison with State-of-the-Art}

MTRNet++ is compared with the previous state-of-the-art one-stage (EnsNet, Pix2Pix) and two-stage (MTRNet) approaches on both Oxford and SCUT datasets. \warn{Note that for fair compaisson, MRTNet++ is provided with empty masks.} In the following section, we will discuss and analyse their results, but before that we briefly introduce their implementation details:
\begin{itemize}
	\item We trained MTRNet++ on both Oxford and SCUT training datasets with the training setup and strategy provided in Section \ref{sec:ts}. MTRNet++ is trained for 200,000 steps on the Oxford training set, and for 85,000 steps on the SCUT training set. For a fair comparison, when MTRNet++ is compared with other state-of-the-art methods, full coarse masks were provided.
	\item  We reimplemented EnsNet with the same settings as MTRNet++ by replacing MTRNet++'s generator with EnsNet's generator. We trained EnsNet on both SCUT and Oxford datasets in the same way that MTRNet++ was trained. We ensure that the reimplemented EnsNet achieves comparable results on the SCUT test set with those reported by \cite{zhang2018ensnet}. As no public EnsNet trained model is available, we used the results of the reimplemented EnsNet for all visual comparisons. 
	\item We tested Pix2Pix and MTRNet using the models trained by \cite{tursun2019mtrnet}. Both were trained for 300,000 steps with a batch size 16. Note that word-level ground-truth bounding box masks were provided for MTRNet.
\end{itemize}

\begin{figure*}[!t]
	\centering
	\includegraphics[width=0.16\linewidth]{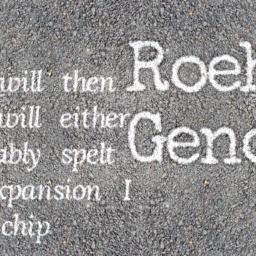}
	\includegraphics[width=0.16\linewidth]{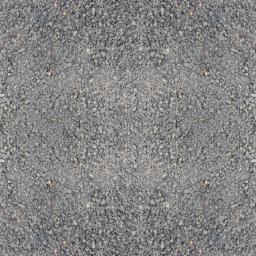}
	\includegraphics[width=0.16\linewidth]{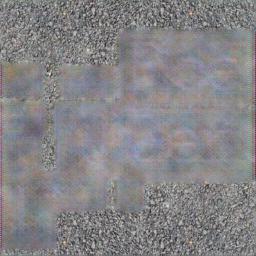}
	\includegraphics[width=0.16\linewidth]{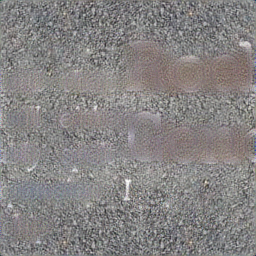}
	\includegraphics[width=0.16\linewidth]{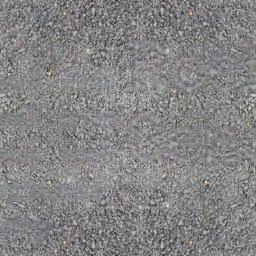}
	\\[0.5ex]
	\includegraphics[width=0.16\linewidth]{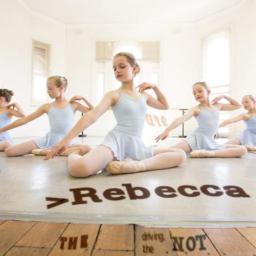}
	\includegraphics[width=0.16\linewidth]{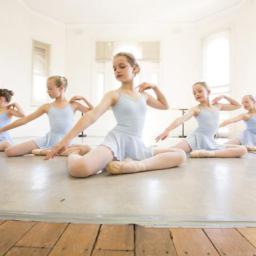}
	\includegraphics[width=0.16\linewidth]{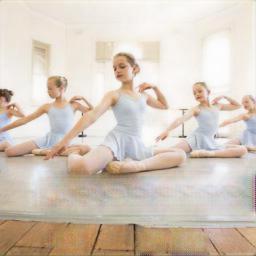}
	\includegraphics[width=0.16\linewidth]{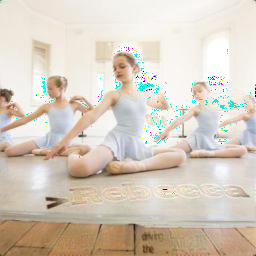}
	\includegraphics[width=0.16\linewidth]{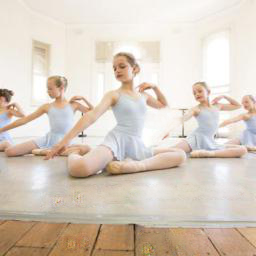}
	\caption{Visual comparison of text removal methods on the Oxford test set. From left to right, input, ground truth, MTRNet, EnsNet and MTRNet++.}
	\label{fig:ox_test}
\end{figure*}

\begin{figure*}[!t]
	\centering
	\includegraphics[width=0.16\linewidth]{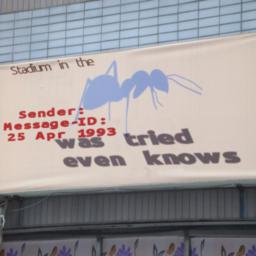}
	\includegraphics[width=0.16\linewidth]{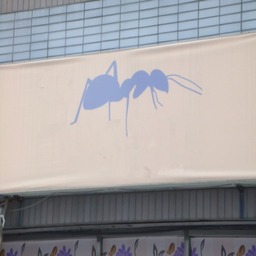}
	\includegraphics[width=0.16\linewidth]{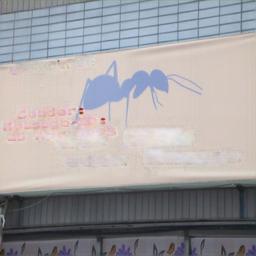}
	\includegraphics[width=0.16\linewidth]{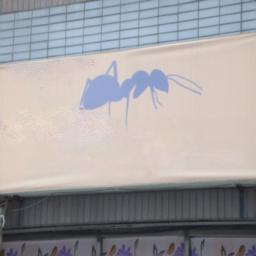}
	\includegraphics[width=0.16\linewidth]{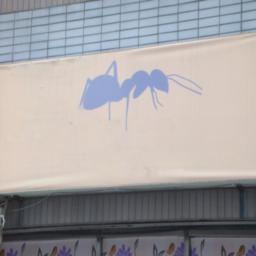}
	\\[0.5ex]
	\includegraphics[width=0.16\linewidth]{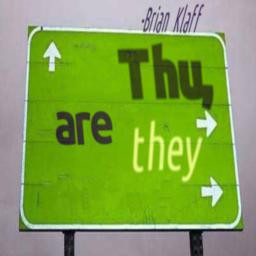}
	\includegraphics[width=0.16\linewidth]{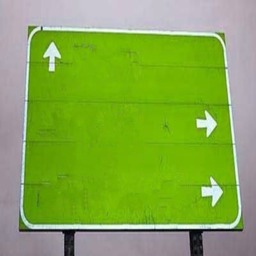}
	\includegraphics[width=0.16\linewidth]{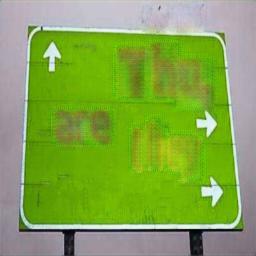}
	\includegraphics[width=0.16\linewidth]{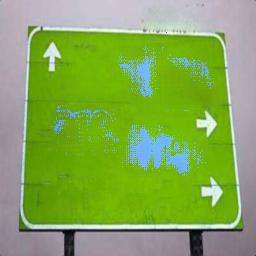}
	\includegraphics[width=0.16\linewidth]{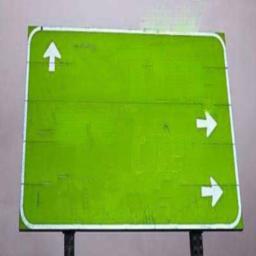}
	\caption{Visual comparison of text removal methods on the SCUT test set. From left to right, input, ground truth, MTRNet, EnsNet and MTRNet++.}
	\label{fig:SCUT_test}
\end{figure*}

Quantitative results of the comparison on Oxford and SCUT test sets are given in Table \ref{tab:ocstr} and \ref{tab:scstr} respectively. Inpainting quality related scores are provided for both Oxford and SCUT datsets, while text detection related scores are only provided for the Oxford dataset as the SCUT dataset's relevant ground-truth is not publicly available yet. Qualitative comparisons are given in Figs. \ref{fig:ox_test} and \ref{fig:SCUT_test}.

MTRNet++ achieved the highest PSNR and SSIM and lowest MSE scores on the Oxford test set by a large margin. It also achieved the second-best performance on the adversarial text detection evaluation with a marginal difference between it and MTRNet, even though MTRNet++ used coarse masks while MTRNet used ground-truth masks. The qualitative results in Fig. \ref{fig:ox_test} show that MTRNet++'s inpainting results are close to the ground-truth, while the results of MTRNet and EnsNet are blurry. What's more, EnsNet's results are also incomplete and partially damaged.

MTRNet++ also achieved comparable inpainting results on the SCUT dataset when compared to the previous state-of-the-art results reported by \cite{zhang2018ensnet}, while it slightly surpassed the reimplemented EnsNet. In Fig. \ref{fig:SCUT_test}, the results of MTRNet++ and EnsNet are close to the ground-truth, while MTRNet's results are blurry and incomplete.

One of the reasons EnsNet's results are comparable to MTRNet++ on a small-scale dataset and not on a large scale dataset is that it lacks good generalisation. We believe that EnsNet memorises the ground-truth in the SCUT dataset as it only has 1,223 ground-truth backgrounds for 8,000 training images. However, MTRNet++ is difficult to overfit even on a small dataset for two reasons: 1. The refined-mask branch output is unique for each input, even though some inputs share the same background, as it predicts text masks on the foreground. 2. The two types of augmentation, padding and filtering, introduced in Section \ref{sec:met} create a new background for each input image.

\warn{MTRNet++ is also a compact and efficient network. It only has 18.7M trainable parameters as indicated in Table \ref{tab:scstr}. Although it has 6.2 million more parameters compared to ENSNet, it has less than one third the parameters of other networks including MTRNet, despite the fact that it has additional mask-prediction and refining branches.}

In summary, MTRNet++ achieved the state-of-the-art results on both Oxford and SCUT datasets. MTRNet++'s results are complete, realistic and stable.


\begin{table*}[!t]
	\small
	\centering
	\bgroup
	\def\arraystretch{1.1}
	\caption{The results of a comparison on the test set of the Oxford Synthetic dataset.}
		\begin{tabular}{l| c | c | c | c | c | c}\hline
			\bf Method    & \bf PSNR   & \bf SSIM(\%)   & \bf MSE (\%) & \bf Precision (\%) & \bf Recall (\%) & \bf F-score\\ \hline
			Original images & - & - & - & 76.41 & 44.27 & 56.06 \\\hline
			Pix2Pix \citep{isola2017image}   & 24.63  & 89.73  & 0.54 & 70.03 & 29.34 & 41.35\\
			MTRNet \citep{tursun2019mtrnet}  & 28.99  & 93.18  & 0.20 & \bf 35.83 & \bf 0.26 & \bf 0.52\\
			EnsNet \citep{zhang2018ensnet}   &  27.42  & 94.37  & 0.21 & 57.25 & 14.34 & 22.94\\\hline
			MTRNet++ & \bf 33.67 &  \bf 98.43  & \bf 0.05 & 50.43 & 1.35 & 2.63\\\hline
		\end{tabular}
	\label{tab:ocstr}
	\egroup
\end{table*}


\begin{table*}[!t]
	\small
	\centering
	\bgroup
	\def\arraystretch{1.1}
	\caption{\warn{Comparison on the test set of the SCUT dataset.}}
	\begin{tabular}{l| c | c | c | c}\hline
		\bf Method    & \bf PSNR   & \bf SSIM(\%)    & \bf MSE (\%) & \bf Parameters\\ \hline
		Pix2Pix \citep{isola2017image}   &  25.60   & 89.86    & 24.56 & 54.4M\\
		STE \citep{nakamura2017scene} & 14.68  & 46.13  & 71.48 & -\\
		MTRNet \citep{tursun2019mtrnet}    & 29.71	 & 94.43  & 0.01 & 54.4M\\
		EnsNet \citep{zhang2018ensnet}    & 37.36  & 96.44   & 0.20 & \bf 12.4M\\
		EnsNet (Reimplemented)    & 34.16  & 98.10   & \bf 0.04 & \bf 12.4M \\
		\cite{zdenek2020erasing} & \bf 37.46  & 93.64   & - & 151.7M\\\hline
		MTRNet++  & 34.55  &  \bf 98.45  & \bf 0.04 & 18.7M \\\hline
	\end{tabular}
	\label{tab:scstr}
	\egroup
\end{table*}

\subsection{Real-life Examples}
\begin{figure*}[!t]
	\centering
	\includegraphics[width=0.2\linewidth]{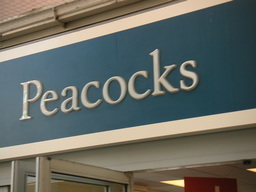}
	\includegraphics[width=0.2\linewidth]{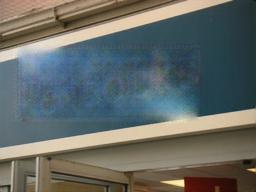}
	\includegraphics[width=0.2\linewidth]{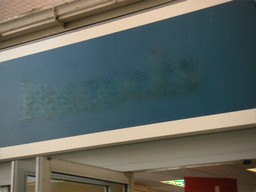}
	\includegraphics[width=0.2\linewidth]{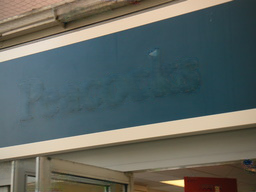}
	\\\vspace{0.05cm}
	\includegraphics[width=0.2\linewidth]{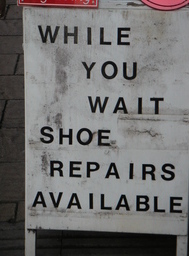}
	\includegraphics[width=0.2\linewidth]{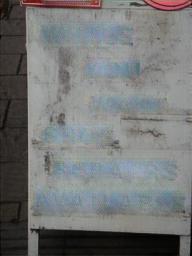}
	\includegraphics[width=0.2\linewidth]{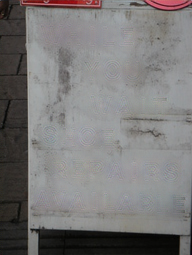}
	\includegraphics[width=0.2\linewidth]{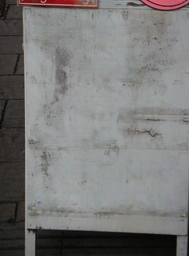}
	\caption{\warn{Examples of real scene text removal. From left to right, input, MTRNet, EnsNet and MTRNet++. Input images are from ICDAR2013 dataset.}}
	\label{fig:ICDAR2013}
\end{figure*}

\warn{Our model is trained using synthetic images, due to the absence of a real scenes dataset for text removal. The domain difference between real and synthetic scenes presents additional challenges to our model. However, our model has successfully removed text from real scene images from the ICDAR2013 dataset as shown in Fig. \ref{fig:ICDAR2013}.}


\subsection{Ablation Studies}



\begin{figure*}[!t]
	\centering
	\includegraphics[width=0.165\linewidth]{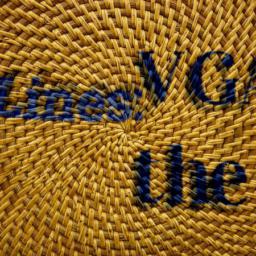}
	\includegraphics[width=0.165\linewidth]{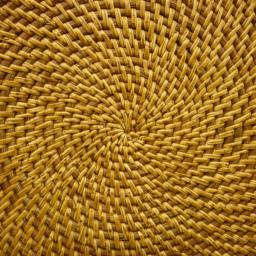}
	\includegraphics[width=0.165\linewidth]{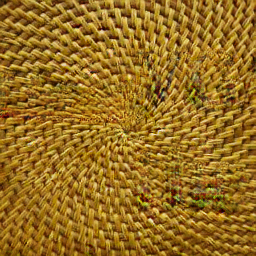}
	\includegraphics[width=0.165\linewidth]{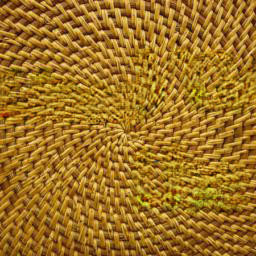}
	\\[0.1ex]
	\includegraphics[width=0.165\linewidth]{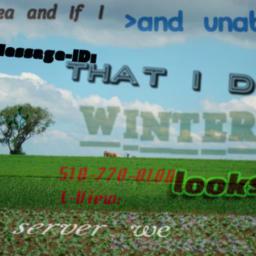}
	\includegraphics[width=0.165\linewidth]{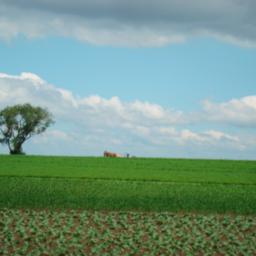}
	\includegraphics[width=0.165\linewidth]{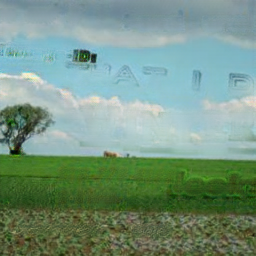}
	\includegraphics[width=0.165\linewidth]{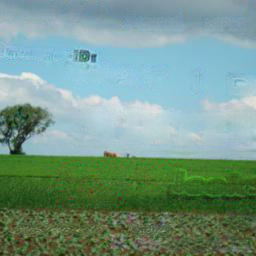}
	\caption{Visual comparison of coarse and fine outputs of MTRNet++. From left to right, input, ground truth, coarse inpaint and fine inpaint.}
	\label{fig:c2f}
\end{figure*}

\noindent{\bf Fine-inpainting Branch} We studied the importance of the fine-inpainting branch by providing composited and predicted coarse and fine inpainting results. Those results for Oxford and SCUT datasets are given in Tables \ref{tab:om} and \ref{tab:sm}. On both datasets the composited fine-inpainting achieved the best results, and has surpassed the predicted coarse-inpainting results by a large margin. In Fig. \ref{fig:c2f}, the composited fine-inpainting has less artifacts than the coarse-inpainted version. Both qualitative and quantitative results show the fine-inpainting branch is befenicial for more realistic inpainting.



\begin{figure}[!h]
	\centering
	\includegraphics[width=\linewidth]{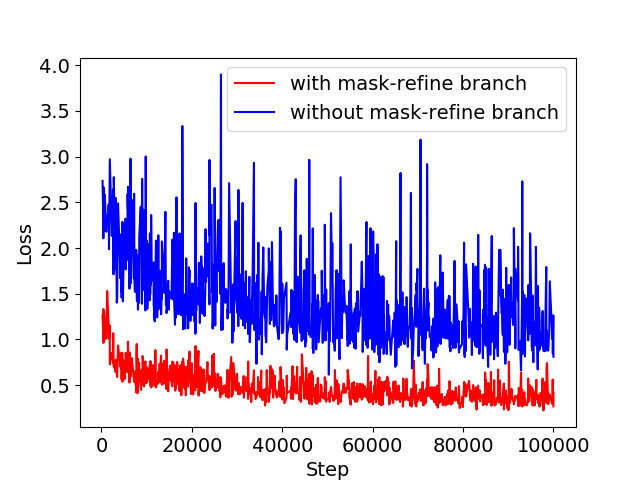}
	\caption{\warn{Training loss curves of MTRNet++ and MTRNet++ without mask-refine branch. MTNet++ converges faster when using the mask-refine branch.}}
	\label{fig:converge}
\end{figure}

\noindent{\bf Mask-Refine Branch} To prove the importance of the mask-refine branch, an MTRNet++ without the mask-refine branch is trained in the same manner as MTRNet++. The related quantitative comparison on the Oxford test set is provided in Table \ref{tab:vmtr}. MTRNet++ surpassed it's no mask-refine branch variant in PSNR, SSIM and MAE scores by a large margin. Moreover, the mask-refine branch is also robust to very coarse masks. For example, MTRNet++'s scores decreases less than its variant when the mask padding is increased in Table \ref{tab:vmtr}. Similar results are also found in Tables \ref{tab:ocstr} and \ref{tab:scstr} where MTRNet++ is less negatively affected by the degree of coarseness of the masks thanks to the mask-refine branch. \warn{Last but not least, with the mask-refine branch, MTRNet++ converges earlier than its no mask counterpart as show in Fig. \ref{fig:converge}. This demonstrates that the mask-refine branch therefore is essential for MTRNet++.}

\noindent{\bf Mask-Refine Loss} We used Tversky loss as the mask-refine loss to ensure a high recall value. From Table \ref{tab:ocstr} and \ref{tab:scstr}, we can see that MTRNet++ achieved very high recall values with high precision values. For example, MTRNet++ achieved at least 97\% recall on both Oxford and Synthetic datasets.

\begin{figure*}[!t]
	\centering
	\includegraphics[width=0.1\linewidth]{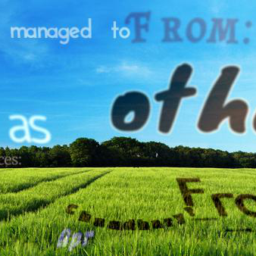}
	\includegraphics[width=0.1\linewidth, frame]{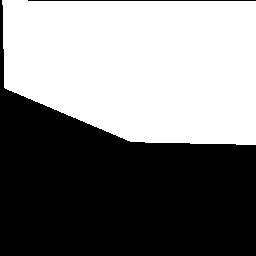}
	\includegraphics[width=0.1\linewidth]{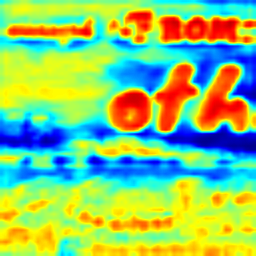}
	\includegraphics[width=0.1\linewidth]{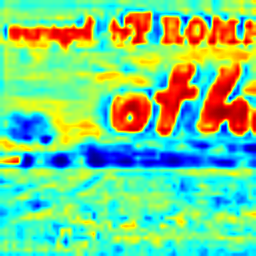}
	\includegraphics[width=0.1\linewidth]{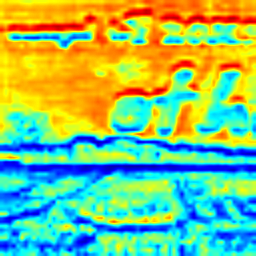}
	\includegraphics[width=0.1\linewidth]{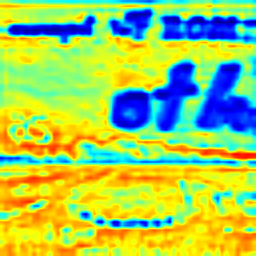}
	\\[0.1ex]
	\includegraphics[width=0.1\linewidth]{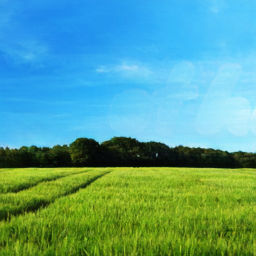}
	\includegraphics[width=0.1\linewidth, frame]{images/attention/1/input_mask}
	\includegraphics[width=0.1\linewidth]{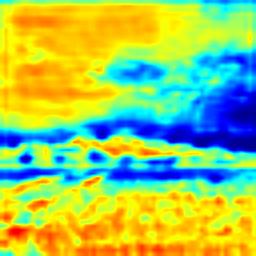}
	\includegraphics[width=0.1\linewidth]{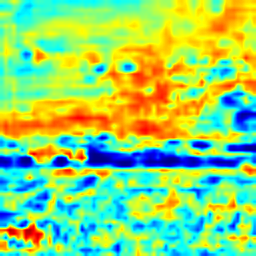}
	\includegraphics[width=0.1\linewidth]{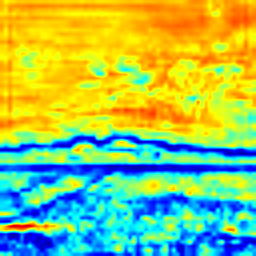}
	\includegraphics[width=0.1\linewidth]{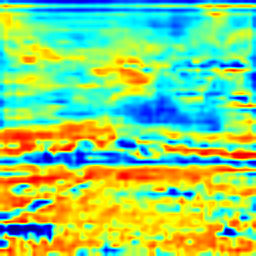}
	\\[0.1ex]
	\includegraphics[width=0.1\linewidth]{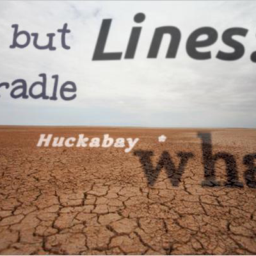}
	\includegraphics[width=0.1\linewidth, frame]{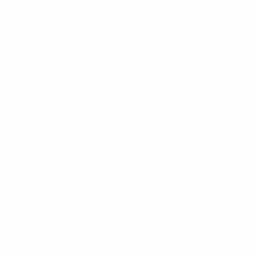}
	\includegraphics[width=0.1\linewidth]{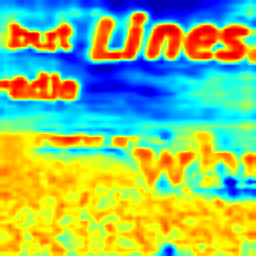}
	\includegraphics[width=0.1\linewidth]{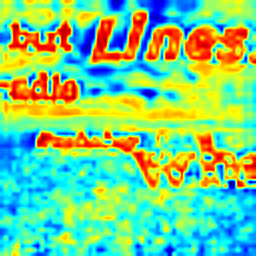}
	\includegraphics[width=0.1\linewidth]{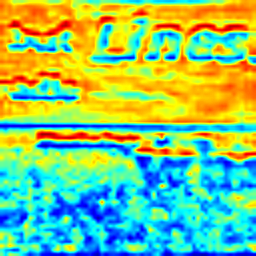}
	\includegraphics[width=0.1\linewidth]{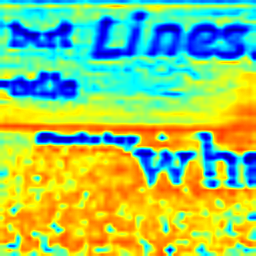}
	\caption{\warn{Example visualisations of the activations of the attention blocks. From left to right, input, mask, attention blocks: A1, A2, A3 and A4. Dark red is the highest attention value, while dark blue is the lowest.}}
	\label{fig:att}
\end{figure*}

\noindent{\bf Attention Block} To test the role of attention blocks in MTRNet++, we trained an MTRNet++ variant without attention blocks. From Table \ref{tab:vmtr}, we can see that MTRNet++'s performance declines when attention blocks are missing. We also visualised the activations of the four attention blocks of the generator of MTRNet++ for randomly selected images in Fig. \ref{fig:att}. Low-level attention blocks, A1 and A2, show high activations on text regions, while high-level attention blocks, A3 and A4, have low activations on text regions. Especially, the attention block A4 assigns the lowest attention to text regions. \warn{The decrease in the attention value of the text region in latter attention blocks is what we expect and desire when introducing attention blocks to the generator. We find low-level attention blocks assign high attention to text-regions. For example, A1 and A2 attention blocks assign high-attention values to regions when text appears as shown in Fig. \ref{fig:att}. We conjecture that low-level attention blocks are used for localizing text regions, while high-level attention blocks suppress features from text regions.  These results show that attention blocks are beneficial for MTRNet++.}

\begin{table*}[]
	\small
	\centering
	\caption{Evaluation of MTRNet++ with various scale masks on the test set of the Oxford dataset. Precision and Recall of refined-masks are calculated for the mask-refine branch. PSNR, SSIM, MAE values are calculated for predicted and composited images from both the coarse-inpainting and fine-inpainting branches.}
	\bgroup
	\def\arraystretch{1.1}
	\begin{tabular}{c | c | c | c | c | c | c |c |c | c | c}\hline
		\multirow{2}{*}{\bf Pad} & \multicolumn{3}{c|}{\bf Mask-Refine} & \multirow{2}{1cm}{\bf Type} &\multicolumn{3}{c|}{\bf Coarse-inpainting}  &    \multicolumn{3}{c}{\bf Fine-inpainting} \\\cline{2-4} \cline{6-11}
		&  \bf Precision (\%)  & \bf Recall (\%) &  \bf F1 & & \bf PSNR   & \bf SSIM (\%)   & \bf MAE (\%) & \bf PSNR   & \bf SSIM (\%)   & \bf MAE (\%) \\\hline
		\multirow{2}{*}{0} &  \multirow{2}{1cm}{82.80} &  \multirow{2}{1cm}{97.74} &  \multirow{2}{1cm}{89.65} & predicted & 29.31   &  95.76   &  5.03 &  32.69  & 97.56 & 3.06 \\\cline{5-11}
		&   &  &  & composited &  35.19  &  98.46   &  0.77 &  35.88  & 98.52 & 0.69 \\\hline
		\multirow{2}{*}{100} & \multirow{2}{1cm}{76.76}  & \multirow{2}{1cm}{97.15} & \multirow{2}{1cm}{85.76} & predicted & 28.95  & 95.66  &5.12  &  31.96  & 97.47 & 3.13   \\\cline{5-11}
		 &   & & & composited & 34.42  &  98.47 & 0.82 &  34.56  & 98.50 &   0.78 \\\hline
		\multirow{2}{*}{256} & \multirow{2}{1cm}{ 72.92} & \multirow{2}{1cm}{ 97.01} & \multirow{2}{1cm}{83.26} & predicted & 28.69   & 95.63  &  5.19 & 31.43  & 97.41 & 3.20 \\\cline{5-11}
		&   &  & & composited & 33.87   & 98.43  &  0.87 & 33.67  & 98.43 & 0.85 \\\hline
	\end{tabular}
	\label{tab:om}
	\egroup
\end{table*}

\begin{table*}[!t]
	\small
	\centering
	\caption{Evaluation of MTRNet++ with various scale masks on the test set of the SCUT dataset. Precision and Recall of refined-masks are calculated for the mask-refine branch. PSNR, SSIM, MAE values are calculated for predicted and composited images from both the coarse-inpainting and fine-inpainting branches.}
	\bgroup
	\def\arraystretch{1.1}
	\begin{tabular}{c | c | c | c | c | c | c |c |c | c | c}\hline
		\multirow{2}{*}{\bf Pad} & \multicolumn{3}{c|}{\bf Mask-Refine} & \multirow{2}{1cm}{\bf Type} &\multicolumn{3}{c|}{\bf Coarse-inpainting}  &    \multicolumn{3}{c}{\bf Fine-inpainting} \\\cline{2-4} \cline{6-11}
		&  \bf Precision (\%)  & \bf Recall (\%) &  \bf F1 & & \bf PSNR   & \bf SSIM (\%)   & \bf MAE (\%) & \bf PSNR   & \bf SSIM (\%)   & \bf MAE (\%) \\\hline
		\multirow{2}{*}{0} &  \multirow{2}{1cm}{88.23} &  \multirow{2}{1cm}{97.97} &  \multirow{2}{1cm}{92.85} & predicted & 30.67   & 95.90  & 3.53 & 33.18  & 97.48 & 2.50 \\\cline{5-11}
		&   &   & & composited &  34.11  & 98.37  &  0.77 & 35.48  & 98.51  &  0.65\\\hline
		\multirow{2}{*}{100} & \multirow{2}{1cm}{81.53}  & \multirow{2}{1cm}{97.92} &  \multirow{2}{1cm}{88.98} & predicted & 30.35  & 95.83  & 3.58  & 32.60   & 97.37 & 2.55   \\\cline{5-11}
		&   &  & & composited &  33.61  &  98.43 & 0.82  & 34.68  &  98.55 & 0.70\\\hline
		\multirow{2}{*}{256} & \multirow{2}{1cm}{80.51} & \multirow{2}{1cm}{97.93} &  \multirow{2}{1cm}{88.37} & predicted & 30.27 & 95.83 & 3.60  & 32.50 & 97.35 & 2.56 \\\cline{5-11}
		&   &   & & composited &  33.50  &  98.42 & 0.83  &  34.55  &  98.54 & 0.71\\\hline
	\end{tabular}
	\label{tab:sm}
	\egroup
\end{table*}






\begin{table*}[!t]
	\centering
	\caption{Comparison of the variants of MTRNet++ on the Oxford Synthetic test. Here, MTRNet++ is compared with its no mask-refine branch and no attention block variants.}
	\bgroup
	\def\arraystretch{1.1}
	\begin{tabular}{c | c | c | c |c |c | c | c |c | c }\hline
		\multirow{2}{*}{\bf Pad} & \multicolumn{3}{c|}{\bf MTRNet++}  & \multicolumn{3}{c|}{\bf MTNet++ without mask-refine}  &    \multicolumn{3}{c}{\bf MTNet++ without attention} \\\cline{2-10}
		& \bf PSNR   & \bf SSIM (\%)   & \bf MAE (\%) & \bf PSNR   & \bf SSIM (\%)   & \bf MAE (\%) & \bf PSNR   & \bf SSIM (\%)   & \bf MAE (\%) \\\hline
	    0   & 35.48  &  98.51 & 0.65  & 26.94   & 93.44   & 6.37 & 33.18  & 98.31  & 0.91\\\hline
	    50  & 34.73  &  98.57 & 0.69  & 25.94   & 92.40   & 6.82 & 32.43  & 98.31  & 0.99\\\hline
	    100 & 34.68  &  98.55 & 0.70  & 25.38   & 91.87   & 7.26 & 32.04  & 98.28  & 1.03\\\hline
	    150 & 34.67  &  98.55 & 0.70  & 24.94   & 91.53   & 7.73 & 31.88  & 98.26  & 1.05\\\hline
	    256 & 34.55  &  98.54 & 0.71  & 24.46   & 91.21   & 8.34 & 31.71  & 98.24  & 1.07\\\hline
	\end{tabular}
	\label{tab:vmtr}
	\egroup
\end{table*}


\subsection{Controllability and Interpretability} 
\begin{figure*}[!ht]
	\centering
	\includegraphics[width=0.165\linewidth]{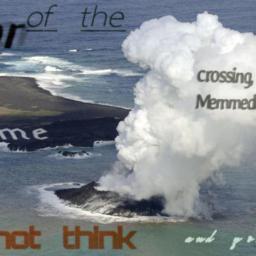}
	\includegraphics[width=0.165\linewidth, frame]{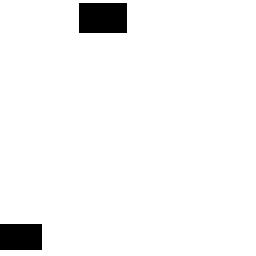}
	\includegraphics[width=0.165\linewidth]{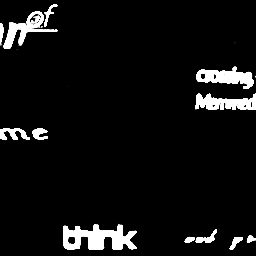}
	\includegraphics[width=0.165\linewidth]{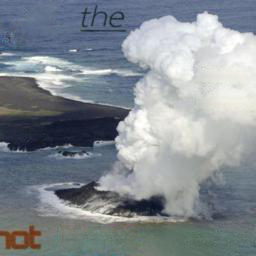}
	\\[0.1ex]
	\includegraphics[width=0.165\linewidth]{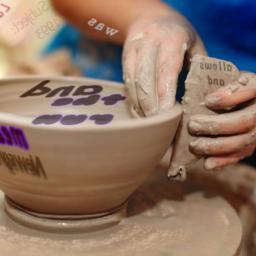}
	\includegraphics[width=0.165\linewidth, frame]{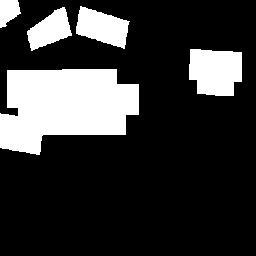}
	\includegraphics[width=0.165\linewidth]{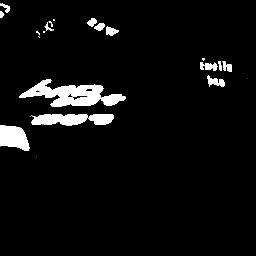}
	\includegraphics[width=0.165\linewidth]{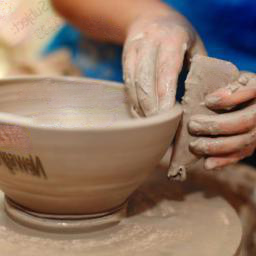}
	\caption{Examples of partial text removal. First and second columns show the inputs; Third and fourth columns are outputs of MTRNet++.}
	\label{fig:ctrl}
\end{figure*}

\noindent{\bf Selective Text Removal} As shown in Fig. \ref{fig:ctrl}, MTRNet++ is able to selectively remove text via designing the masks.   MTRNet++ only erases text under the given masks, and only generates refined masks for text that needs to be removed.

\begin{figure*}[!t]
	\centering
	\includegraphics[width=0.16\linewidth]{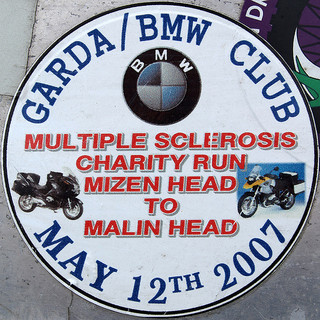}
	\includegraphics[width=0.16\linewidth, frame]{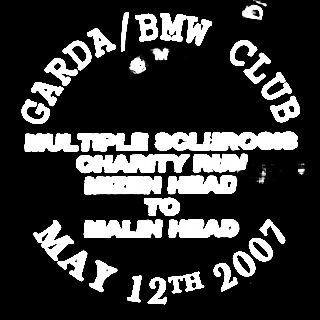}
	\includegraphics[width=0.16\linewidth]{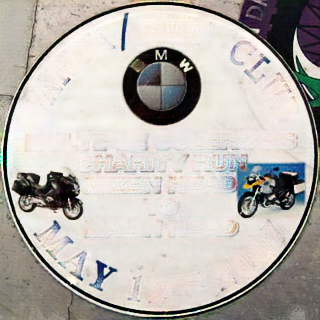}
	\includegraphics[width=0.16\linewidth]{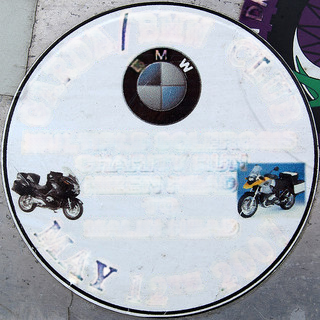}
	\\[0.1ex]
	\includegraphics[width=0.16\linewidth]{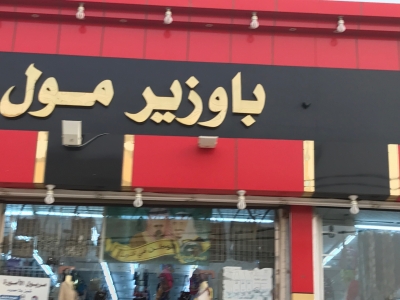}
	\includegraphics[width=0.16\linewidth, frame]{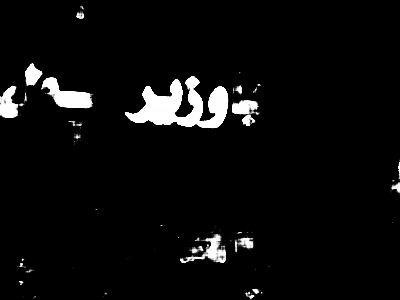}
	\includegraphics[width=0.16\linewidth]{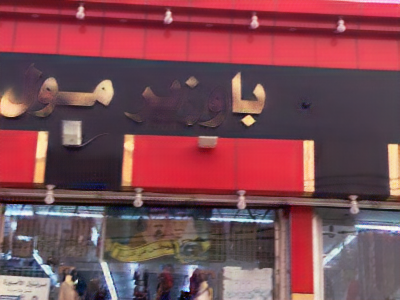}
	\includegraphics[width=0.16\linewidth]{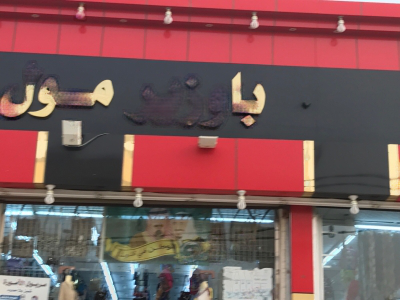}
	\caption{Examples of interpretability. From left to right, input, predicted mask and predicted image.}
	\label{fig:inter}
\end{figure*}

\noindent{\bf Interpretability} Interpretability is important for diagnosing failure cases. MTRNet++'s intermediate results help analysing failure cases. For example, two failure cases are given in Fig. \ref{fig:inter}. The reason for the first case (top row) is related to inpainting branches since the predicted mask is accurate. In comparison, the second case (bottom row) is related to mask-refine branch as the mask is not complete.

\subsection{Generalisation to General Object Removal}
\begin{figure*}[!t]
	\centering
	\includegraphics[width=0.13\linewidth]{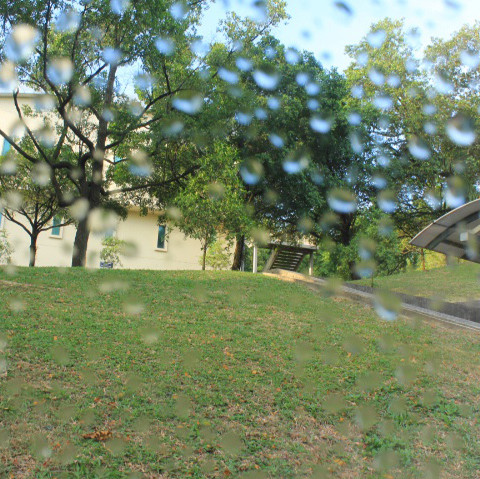}
	\includegraphics[width=0.13\linewidth, frame]{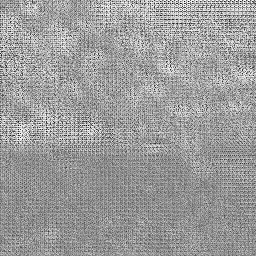}
	\includegraphics[width=0.13\linewidth]{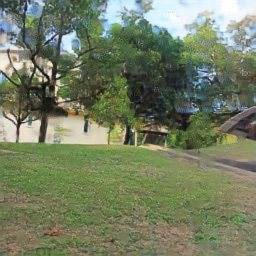}
	\includegraphics[width=0.13\linewidth]{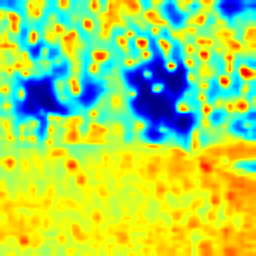}
	\includegraphics[width=0.13\linewidth]{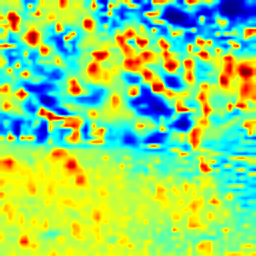}
	\includegraphics[width=0.13\linewidth]{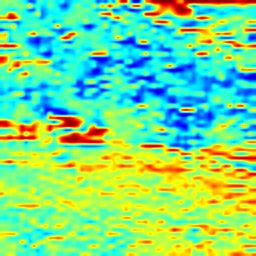}
	\includegraphics[width=0.13\linewidth]{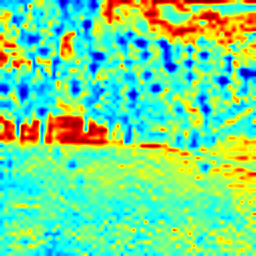}
	\caption{\warn{Example output for the raindrop removal task. From left to right: input, predicted mask and predicted image, and the output of the four attention blocks A1-A4. Dark red is the highest attention value, while dark blue is the lowest. The MTRNet++ failed to predict a valid refined mask for raindrops as it was not trained with mask-refine loss.}}
	\label{fig:rd}
\end{figure*}

\warn{Our model is proposed for text removal tasks, however it can be applied to other objects as well. We tested our method on the Raindrop dataset of \cite{qian2018attentive} as it is a very challenging object removal task. The Raindrop dataset has a high occlusion rate, fewer training samples (861) and no ground-truth masks. We trained MTRNet++ without mask-refine loss due to the absence of ground-truth raindrop masks. We quantitatively compared MTRNet++ with methods reported by \cite{qian2018attentive}, and results are shown in Table \ref{tab:rd}. We have achieved the state-of-the-art SSIM score, though achieve a lower PSNR. We also visualized output of the MTRNet++ and it's attention blocks. As shown in Fig. \ref{fig:rd}, MTRNet++ has removed the raindrops, but failed to generate a valid refine-mask for raindrops. However, the attention blocks have still learned to localize raindrops (A1 and A2 blocks) and suppress raindrops' features (A3 and A4 blocks). In conclusion, our model can be easily extended to general object removal tasks.}

\begin{table}[!t]
	\centering
	\bgroup
	\def\arraystretch{1.1}
	\caption{\warn{Evaluation results for the test set of the Raindrop dataset.}}
	\begin{tabular}{l| c | c }\hline
		\bf Method    & \bf PSNR   & \bf SSIM(\%) \\ \hline
		\cite{eigen2013restoring}  & 28.59   & 67.26  \\
		Pix2Pix (\cite{isola2017image}) & 30.14  & 82.99   \\
		Attentive GAN (\cite{qian2018attentive}) & \textbf {31.57}      & 90.23     \\\hline
		MTRNet++ without  & \multirow{2}{*}{24.46 }   & \multirow{2}{*}{\textbf {90.25}} \\
		mask-refine loss (ours) &    &  \\\hline
	\end{tabular}
	\label{tab:rd}
	\egroup
\end{table}

\section{Conclusion}
In this work, a one-stage mask-based conditional generative adversarial network, MTRNet++, is proposed for real-scene text removal. It is self-complete, controllable and interpretable. It shows state-of-the-art quantitative results on both the Oxford and SCUT test datasets without user-provided text masks. Visual results also show MTRNet++ generates realistic inpainting for text regions. Related ablation studies show that proposed multi-branch generator is essential for state-of-the-art performance. Moreover, MTRNet++ is easily trainable in an end-to-end way. MTRNet++ converges on a large-scale dataset within an epoch. MTRNet++ also shows its controllability and interpretability.

\section*{Acknowledgements}
This research was supported by an Advance Queensland Research Fellowship Award.

\bibliographystyle{model2-names}
\bibliography{egbib}

\begin{thebibliography}{33}
\expandafter\ifx\csname natexlab\endcsname\relax\def\natexlab#1{#1}\fi
\providecommand{\url}[1]{\texttt{#1}}
\providecommand{\href}[2]{#2}
\providecommand{\path}[1]{#1}
\providecommand{\DOIprefix}{doi:}
\providecommand{\ArXivprefix}{arXiv:}
\providecommand{\URLprefix}{URL: }
\providecommand{\Pubmedprefix}{pmid:}
\providecommand{\doi}[1]{\href{http://dx.doi.org/#1}{\path{#1}}}
\providecommand{\Pubmed}[1]{\href{pmid:#1}{\path{#1}}}
\providecommand{\bibinfo}[2]{#2}
\ifx\xfnm\relax \def\xfnm[#1]{\unskip,\space#1}\fi
\bibitem[{Baek et~al.(2019)Baek, Lee, Han, Yun and Lee}]{baek2019character}
\bibinfo{author}{Baek, Y.}, \bibinfo{author}{Lee, B.}, \bibinfo{author}{Han,
  D.}, \bibinfo{author}{Yun, S.}, \bibinfo{author}{Lee, H.},
  \bibinfo{year}{2019}.
\newblock \bibinfo{title}{Character region awareness for text detection}, in:
  \bibinfo{booktitle}{Proceedings of the IEEE Conference on Computer Vision and
  Pattern Recognition}, pp. \bibinfo{pages}{9365--9374}.
\bibitem[{Eigen et~al.(2013)Eigen, Krishnan and Fergus}]{eigen2013restoring}
\bibinfo{author}{Eigen, D.}, \bibinfo{author}{Krishnan, D.},
  \bibinfo{author}{Fergus, R.}, \bibinfo{year}{2013}.
\newblock \bibinfo{title}{Restoring an image taken through a window covered
  with dirt or rain}, in: \bibinfo{booktitle}{Proceedings of the IEEE
  international conference on computer vision}, pp. \bibinfo{pages}{633--640}.
\bibitem[{Gupta et~al.(2016)Gupta, Vedaldi and Zisserman}]{gupta2016synthetic}
\bibinfo{author}{Gupta, A.}, \bibinfo{author}{Vedaldi, A.},
  \bibinfo{author}{Zisserman, A.}, \bibinfo{year}{2016}.
\newblock \bibinfo{title}{Synthetic data for text localisation in natural
  images}, in: \bibinfo{booktitle}{Proc. of CVPR}.
\bibitem[{He et~al.(2016)He, Zhang, Ren and Sun}]{he2016deep}
\bibinfo{author}{He, K.}, \bibinfo{author}{Zhang, X.}, \bibinfo{author}{Ren,
  S.}, \bibinfo{author}{Sun, J.}, \bibinfo{year}{2016}.
\newblock \bibinfo{title}{Deep residual learning for image recognition}, in:
  \bibinfo{booktitle}{Proceedings of the IEEE conference on computer vision and
  pattern recognition}, pp. \bibinfo{pages}{770--778}.
\bibitem[{Iizuka et~al.(2017)Iizuka, Simo-Serra and
  Ishikawa}]{iizuka2017globally}
\bibinfo{author}{Iizuka, S.}, \bibinfo{author}{Simo-Serra, E.},
  \bibinfo{author}{Ishikawa, H.}, \bibinfo{year}{2017}.
\newblock \bibinfo{title}{Globally and locally consistent image completion}.
\newblock \bibinfo{journal}{ToG} .
\bibitem[{Isola et~al.(2017)Isola, Zhu, Zhou and Efros}]{isola2017image}
\bibinfo{author}{Isola, P.}, \bibinfo{author}{Zhu, J.Y.},
  \bibinfo{author}{Zhou, T.}, \bibinfo{author}{Efros, A.A.},
  \bibinfo{year}{2017}.
\newblock \bibinfo{title}{Image-to-image translation with conditional
  adversarial networks}.
\newblock \bibinfo{journal}{CVPR} .
\bibitem[{Jo and Park(2019)}]{jo2019sc}
\bibinfo{author}{Jo, Y.}, \bibinfo{author}{Park, J.}, \bibinfo{year}{2019}.
\newblock \bibinfo{title}{Sc-fegan: Face editing generative adversarial network
  with user's sketch and color}.
\newblock \bibinfo{journal}{arXiv preprint arXiv:1902.06838} .
\bibitem[{Johnson et~al.(2016)Johnson, Alahi and
  Fei-Fei}]{johnson2016perceptual}
\bibinfo{author}{Johnson, J.}, \bibinfo{author}{Alahi, A.},
  \bibinfo{author}{Fei-Fei, L.}, \bibinfo{year}{2016}.
\newblock \bibinfo{title}{Perceptual losses for real-time style transfer and
  super-resolution}, in: \bibinfo{booktitle}{ECCV}.
\bibitem[{Karatzas et~al.(2013)Karatzas, Shafait, Uchida, Iwamura, i~Bigorda,
  Mestre, Mas, Mota, Almazan and De~Las~Heras}]{karatzas2013icdar}
\bibinfo{author}{Karatzas, D.}, \bibinfo{author}{Shafait, F.},
  \bibinfo{author}{Uchida, S.}, \bibinfo{author}{Iwamura, M.},
  \bibinfo{author}{i~Bigorda, L.G.}, \bibinfo{author}{Mestre, S.R.},
  \bibinfo{author}{Mas, J.}, \bibinfo{author}{Mota, D.F.},
  \bibinfo{author}{Almazan, J.A.}, \bibinfo{author}{De~Las~Heras, L.P.},
  \bibinfo{year}{2013}.
\newblock \bibinfo{title}{Icdar 2013 robust reading competition}, in:
  \bibinfo{booktitle}{ICDAR}.
\bibitem[{Khodadadi and Behrad(2012)}]{khodadadi2012text}
\bibinfo{author}{Khodadadi, M.}, \bibinfo{author}{Behrad, A.},
  \bibinfo{year}{2012}.
\newblock \bibinfo{title}{Text localization, extraction and inpainting in color
  images}, in: \bibinfo{booktitle}{2012 20th Iranian Conf. on Electrical
  Engineering (ICEE)}, pp. \bibinfo{pages}{1035--1040}.
\bibitem[{Liu et~al.(2018)Liu, Reda, Shih, Wang, Tao and
  Catanzaro}]{liu2018image}
\bibinfo{author}{Liu, G.}, \bibinfo{author}{Reda, F.A.}, \bibinfo{author}{Shih,
  K.J.}, \bibinfo{author}{Wang, T.C.}, \bibinfo{author}{Tao, A.},
  \bibinfo{author}{Catanzaro, B.}, \bibinfo{year}{2018}.
\newblock \bibinfo{title}{Image inpainting for irregular holes using partial
  convolutions}, in: \bibinfo{booktitle}{Proc. of ECCV}, pp.
  \bibinfo{pages}{85--100}.
\bibitem[{Ma et~al.(2019)Ma, Liu, Bai, Wang, He and Liu}]{ma2019coarse}
\bibinfo{author}{Ma, Y.}, \bibinfo{author}{Liu, X.}, \bibinfo{author}{Bai, S.},
  \bibinfo{author}{Wang, L.}, \bibinfo{author}{He, D.}, \bibinfo{author}{Liu,
  A.}, \bibinfo{year}{2019}.
\newblock \bibinfo{title}{Coarse-to-fine image inpainting via region-wise
  convolutions and non-local correlation}, in: \bibinfo{booktitle}{Proceedings
  of the 28th International Joint Conference on Artificial Intelligence},
  \bibinfo{organization}{AAAI Press}. pp. \bibinfo{pages}{3123--3129}.
\bibitem[{Maas et~al.(2013)Maas, Hannun and Ng}]{maas2013rectifier}
\bibinfo{author}{Maas, A.L.}, \bibinfo{author}{Hannun, A.Y.},
  \bibinfo{author}{Ng, A.Y.}, \bibinfo{year}{2013}.
\newblock \bibinfo{title}{Rectifier nonlinearities improve neural network
  acoustic models}, in: \bibinfo{booktitle}{Proc. icml}, p.~\bibinfo{pages}{3}.
\bibitem[{Mao et~al.(2016)Mao, Shen and Yang}]{mao2016image}
\bibinfo{author}{Mao, X.J.}, \bibinfo{author}{Shen, C.}, \bibinfo{author}{Yang,
  Y.B.}, \bibinfo{year}{2016}.
\newblock \bibinfo{title}{Image restoration using convolutional auto-encoders
  with symmetric skip connections}.
\newblock \bibinfo{journal}{arXiv preprint arXiv:1606.08921} .
\bibitem[{Miyato et~al.(2018)Miyato, Kataoka, Koyama and
  Yoshida}]{miyato2018spectral}
\bibinfo{author}{Miyato, T.}, \bibinfo{author}{Kataoka, T.},
  \bibinfo{author}{Koyama, M.}, \bibinfo{author}{Yoshida, Y.},
  \bibinfo{year}{2018}.
\newblock \bibinfo{title}{Spectral normalization for generative adversarial
  networks}.
\newblock \bibinfo{journal}{arXiv preprint arXiv:1802.05957} .
\bibitem[{Nakamura et~al.(2017)Nakamura, Zhu, Yanai and
  Uchida}]{nakamura2017scene}
\bibinfo{author}{Nakamura, T.}, \bibinfo{author}{Zhu, A.},
  \bibinfo{author}{Yanai, K.}, \bibinfo{author}{Uchida, S.},
  \bibinfo{year}{2017}.
\newblock \bibinfo{title}{Scene text eraser}, in: \bibinfo{booktitle}{ICDAR},
  pp. \bibinfo{pages}{832--837}.
\bibitem[{Nayef et~al.(2017)Nayef, Yin, Bizid, Choi, Feng, Karatzas, Luo, Pal,
  Rigaud, Chazalon et~al.}]{nayef2017icdar2017}
\bibinfo{author}{Nayef, N.}, \bibinfo{author}{Yin, F.}, \bibinfo{author}{Bizid,
  I.}, \bibinfo{author}{Choi, H.}, \bibinfo{author}{Feng, Y.},
  \bibinfo{author}{Karatzas, D.}, \bibinfo{author}{Luo, Z.},
  \bibinfo{author}{Pal, U.}, \bibinfo{author}{Rigaud, C.},
  \bibinfo{author}{Chazalon, J.}, et~al., \bibinfo{year}{2017}.
\newblock \bibinfo{title}{Icdar2017 robust reading challenge on multi-lingual
  scene text detection and script identification-rrc-mlt}, in:
  \bibinfo{booktitle}{ICDAR}, pp. \bibinfo{pages}{1454--1459}.
\bibitem[{Nazeri et~al.(2019)Nazeri, Ng, Joseph, Qureshi and
  Ebrahimi}]{nazeri2019edgeconnect}
\bibinfo{author}{Nazeri, K.}, \bibinfo{author}{Ng, E.},
  \bibinfo{author}{Joseph, T.}, \bibinfo{author}{Qureshi, F.},
  \bibinfo{author}{Ebrahimi, M.}, \bibinfo{year}{2019}.
\newblock \bibinfo{title}{Edgeconnect: Generative image inpainting with
  adversarial edge learning}.
\newblock \bibinfo{journal}{arXiv preprint arXiv:1901.00212} .
\bibitem[{Qian et~al.(2018)Qian, Tan, Yang, Su and Liu}]{qian2018attentive}
\bibinfo{author}{Qian, R.}, \bibinfo{author}{Tan, R.T.}, \bibinfo{author}{Yang,
  W.}, \bibinfo{author}{Su, J.}, \bibinfo{author}{Liu, J.},
  \bibinfo{year}{2018}.
\newblock \bibinfo{title}{Attentive generative adversarial network for raindrop
  removal from a single image}, in: \bibinfo{booktitle}{Proceedings of the IEEE
  conference on computer vision and pattern recognition}, pp.
  \bibinfo{pages}{2482--2491}.
\bibitem[{Salehi et~al.(2017)Salehi, Erdogmus and
  Gholipour}]{salehi2017tversky}
\bibinfo{author}{Salehi, S.S.M.}, \bibinfo{author}{Erdogmus, D.},
  \bibinfo{author}{Gholipour, A.}, \bibinfo{year}{2017}.
\newblock \bibinfo{title}{Tversky loss function for image segmentation using 3d
  fully convolutional deep networks}, in: \bibinfo{booktitle}{Int. Workshop on
  MLMI}, \bibinfo{organization}{Springer}.
\bibitem[{Simonyan and Zisserman(2014)}]{simonyan2014very}
\bibinfo{author}{Simonyan, K.}, \bibinfo{author}{Zisserman, A.},
  \bibinfo{year}{2014}.
\newblock \bibinfo{title}{Very deep convolutional networks for large-scale
  image recognition}.
\newblock \bibinfo{journal}{arXiv preprint arXiv:1409.1556} .
\bibitem[{Tursun et~al.(2019a)Tursun, Denman, Sivapalan, Sridharan, Fookes and
  Mau}]{tursun2019component}
\bibinfo{author}{Tursun, O.}, \bibinfo{author}{Denman, S.},
  \bibinfo{author}{Sivapalan, S.}, \bibinfo{author}{Sridharan, S.},
  \bibinfo{author}{Fookes, C.}, \bibinfo{author}{Mau, S.},
  \bibinfo{year}{2019}a.
\newblock \bibinfo{title}{Component-based attention for large-scale trademark
  retrieval}.
\newblock \bibinfo{journal}{IEEE Transactions on Information Forensics and
  Security} .
\bibitem[{Tursun et~al.(2019b)Tursun, Zeng, Denman, Sivipalan, Sridharan and
  Fookes}]{tursun2019mtrnet}
\bibinfo{author}{Tursun, O.}, \bibinfo{author}{Zeng, R.},
  \bibinfo{author}{Denman, S.}, \bibinfo{author}{Sivipalan, S.},
  \bibinfo{author}{Sridharan, S.}, \bibinfo{author}{Fookes, C.},
  \bibinfo{year}{2019}b.
\newblock \bibinfo{title}{Mtrnet: A generic scene text eraser}, in:
  \bibinfo{booktitle}{International Conference on Document Analysis and
  Recognition (ICDAR)}, \bibinfo{organization}{IEEE}. pp.
  \bibinfo{pages}{39--44}.
\bibitem[{Ulyanov et~al.(2017)Ulyanov, Vedaldi and
  Lempitsky}]{ulyanov2017improved}
\bibinfo{author}{Ulyanov, D.}, \bibinfo{author}{Vedaldi, A.},
  \bibinfo{author}{Lempitsky, V.}, \bibinfo{year}{2017}.
\newblock \bibinfo{title}{Improved texture networks: Maximizing quality and
  diversity in feed-forward stylization and texture synthesis}, in:
  \bibinfo{booktitle}{Proc. of CVPR}, pp. \bibinfo{pages}{6924--6932}.
\bibitem[{Wagh and Patil(2015)}]{wagh2015text}
\bibinfo{author}{Wagh, P.D.}, \bibinfo{author}{Patil, D.},
  \bibinfo{year}{2015}.
\newblock \bibinfo{title}{Text detection and removal from image using
  inpainting with smoothing}, in: \bibinfo{booktitle}{2015 Int. Conf. on
  Pervasive Computing (ICPC)}, pp. \bibinfo{pages}{1--4}.
\bibitem[{Wolf et~al.(2014)Wolf, Lombardi, Mille, Celiktutan, Jiu, Dogan, Eren,
  Baccouche, Dellandr{\'e}a, Bichot et~al.}]{wolf2014evaluation}
\bibinfo{author}{Wolf, C.}, \bibinfo{author}{Lombardi, E.},
  \bibinfo{author}{Mille, J.}, \bibinfo{author}{Celiktutan, O.},
  \bibinfo{author}{Jiu, M.}, \bibinfo{author}{Dogan, E.},
  \bibinfo{author}{Eren, G.}, \bibinfo{author}{Baccouche, M.},
  \bibinfo{author}{Dellandr{\'e}a, E.}, \bibinfo{author}{Bichot, C.E.}, et~al.,
  \bibinfo{year}{2014}.
\newblock \bibinfo{title}{Evaluation of video activity localizations
  integrating quality and quantity measurements}.
\newblock \bibinfo{journal}{Computer Vision and Image Understanding (CVIU)}
  \bibinfo{volume}{127}, \bibinfo{pages}{14--30}.
\bibitem[{Yang et~al.(2017)Yang, Lu, Lin, Shechtman, Wang and
  Li}]{yang2017high}
\bibinfo{author}{Yang, C.}, \bibinfo{author}{Lu, X.}, \bibinfo{author}{Lin,
  Z.}, \bibinfo{author}{Shechtman, E.}, \bibinfo{author}{Wang, O.},
  \bibinfo{author}{Li, H.}, \bibinfo{year}{2017}.
\newblock \bibinfo{title}{High-resolution image inpainting using multi-scale
  neural patch synthesis}, in: \bibinfo{booktitle}{Proc. of the CVPR}.
\bibitem[{Yu and Koltun(2015)}]{yu2015multi}
\bibinfo{author}{Yu, F.}, \bibinfo{author}{Koltun, V.}, \bibinfo{year}{2015}.
\newblock \bibinfo{title}{Multi-scale context aggregation by dilated
  convolutions}.
\newblock \bibinfo{journal}{arXiv preprint arXiv:1511.07122} .
\bibitem[{Yu and Koltun(2016)}]{YuKoltun2016}
\bibinfo{author}{Yu, F.}, \bibinfo{author}{Koltun, V.}, \bibinfo{year}{2016}.
\newblock \bibinfo{title}{Multi-scale context aggregation by dilated
  convolutions}, in: \bibinfo{booktitle}{ICLR}.
\bibitem[{Yu et~al.(2018a)Yu, Lin, Yang, Shen, Lu and Huang}]{yu2018free}
\bibinfo{author}{Yu, J.}, \bibinfo{author}{Lin, Z.}, \bibinfo{author}{Yang,
  J.}, \bibinfo{author}{Shen, X.}, \bibinfo{author}{Lu, X.},
  \bibinfo{author}{Huang, T.S.}, \bibinfo{year}{2018}a.
\newblock \bibinfo{title}{Free-form image inpainting with gated convolution}.
\newblock \bibinfo{journal}{arXiv preprint arXiv:1806.03589} .
\bibitem[{Yu et~al.(2018b)Yu, Lin, Yang, Shen, Lu and Huang}]{yu2018generative}
\bibinfo{author}{Yu, J.}, \bibinfo{author}{Lin, Z.}, \bibinfo{author}{Yang,
  J.}, \bibinfo{author}{Shen, X.}, \bibinfo{author}{Lu, X.},
  \bibinfo{author}{Huang, T.S.}, \bibinfo{year}{2018}b.
\newblock \bibinfo{title}{Generative image inpainting with contextual
  attention}.
\newblock \bibinfo{journal}{arXiv preprint arXiv:1801.07892} .
\bibitem[{Zdenek and Nakayama(2020)}]{zdenek2020erasing}
\bibinfo{author}{Zdenek, J.}, \bibinfo{author}{Nakayama, H.},
  \bibinfo{year}{2020}.
\newblock \bibinfo{title}{Erasing scene text with weak supervision}, in:
  \bibinfo{booktitle}{The IEEE Winter Conference on Applications of Computer
  Vision}, pp. \bibinfo{pages}{2238--2246}.
\bibitem[{Zhang et~al.(2018)Zhang, Liu, Jin, Huang and Lai}]{zhang2018ensnet}
\bibinfo{author}{Zhang, S.}, \bibinfo{author}{Liu, Y.}, \bibinfo{author}{Jin,
  L.}, \bibinfo{author}{Huang, Y.}, \bibinfo{author}{Lai, S.},
  \bibinfo{year}{2018}.
\newblock \bibinfo{title}{Ensnet: Ensconce text in the wild}.
\newblock \bibinfo{journal}{arXiv preprint arXiv:1812.00723} .

\end{thebibliography}

\appendix
\section{Generator and Discriminator Architectures}
\label{ap:ap1}
Notations described in Table \ref{tab:not} will be used for describing the structure of generator and discriminator. The generator architecture is given in Fig. \ref{fig:model}. The generator repeatedly uses several component blocks whose structures are show in Table \ref{tab:ginfo}. The structure of the discriminator is as follow:

\noindent {\fontfamily{qcr}\selectfont
	64c4p1s2-SP-LRL, MC-128c4p1s2-SP-LRL, MC-256c4p1s2-SP-LRL, MC-512c4p1s1-SP-LRL, MC-1c4p1s1-SP}
%
%
\begin{table}[!h]
	\small
	\centering
	\caption{The notation and their explanation used for describing the network architecture.}
	\bgroup
	\def\arraystretch{1.1}
	\begin{tabular}{l | p{3cm} | p{4cm}} \hline
	\bf Not. & \bf Translation  & \bf Usage \\\hline
	\fontfamily{qcr}\selectfont c  &  2D convolution & {\fontfamily{qcr}\selectfont mcn} is $n\times n$ convolution with $m$ filters.\\\hline
	\fontfamily{qcr}\selectfont s  &  2D convolution stride & {\fontfamily{qcr}\selectfont sn} is a stride with step size $n$. The default step size is 1. \\\hline
	\fontfamily{qcr}\selectfont p  &  2D convolution padding & {\fontfamily{qcr}\selectfont pn} is a padding with size $n$. The default padding size is 0. \\\hline
	\fontfamily{qcr}\selectfont d  & 2D convolution dilation \cite{yu2015multi} & {\fontfamily{qcr}\selectfont dn} is a dilation with size $n$. The default dilation size is 1. \\\hline
	\fontfamily{qcr}\selectfont RP   &  Reflection padding &  {\fontfamily{qcr}\selectfont RPn} is a reflection padding with the size of $n$. \\\hline
	\fontfamily{qcr}\selectfont IN & Instance normalisation \cite{ulyanov2017improved} & \\\hline
	\fontfamily{qcr}\selectfont SP & Spectral normalisation \cite{miyato2018spectral} & \\\hline
	\fontfamily{qcr}\selectfont RC  & Residual connection \cite{he2016deep} \\\hline
	\fontfamily{qcr}\selectfont MC & Mask concatenation & the {\fontfamily{qcr}\selectfont MC} operation concatenates the input mask to the previous layer output. The mask is resized to the previous layer output size via nearest neighbour resize. \\\hline
	\fontfamily{qcr}\selectfont RL & Relu & \\\hline 
	\fontfamily{qcr}\selectfont S  & Sigmoid  & \\\hline
	\fontfamily{qcr}\selectfont LRL & Leaky relu \citep{maas2013rectifier} &  LeakyReLU is employed with slope 0.2. \\\hline
	\end{tabular}
	\label{tab:not}
	\egroup
\end{table}
\begin{table}[!h]
	\small
	\centering
	\caption{The detailed structure of components of the generator.}
	\bgroup
	\def\arraystretch{1.1}
	\begin{tabular}{l | p{5cm}} \hline
		\bf Component & \bf Structure \\\hline
		\multirow{3}{2cm}{\bf Front-end} & \fontfamily{qcr}\selectfont RP3-64c4p1s2-IN-RL, \\
		     & \fontfamily{qcr}\selectfont 128c4p1s2-IN-RL, \\
		     & \fontfamily{qcr}\selectfont 128c4p1s2-IN-RL \\\hline
		\multirow{2}{3cm}{\bf Back-end \\(Inpaint branch)}  &  \fontfamily{qcr}\selectfont 256u4p1s2-IN-RL, \\
		& \fontfamily{qcr}\selectfont 128c4p1s2-IN-RL, \fontfamily{qcr}\selectfont RP3-3c7\\\hline
		\multirow{2}{3cm}{\bf Back-end (Mask-refine branch)} &  \fontfamily{qcr}\selectfont 256u4p1s2-IN-RL, \\
		& \fontfamily{qcr}\selectfont 128c4p1s2-IN-RL, \fontfamily{qcr}\selectfont RP3-1c7 \\\hline
		\multirow{2}{2cm}{\bf Residual block} & \fontfamily{qcr}\selectfont RP2-256c3d2-IN-RL, RP1-256c3-IN-RC \\\hline
		\multirow{2}{3cm}{\bf Attention block} & \fontfamily{qcr}\selectfont RP1-128c3-IN-RL, \\
		& \fontfamily{qcr}\selectfont RP1-1c3-IN-S\\\hline
	\end{tabular}
	\label{tab:ginfo}
	\egroup
\end{table}
\end{document}